\documentclass[nohyperref]{article}

\usepackage{microtype}
\usepackage{graphicx}
\usepackage{subfigure}
\usepackage{booktabs} 
\UseRawInputEncoding
\usepackage{hyperref}

\usepackage[accepted]{icml2022}

\usepackage{amsmath}
\usepackage{amssymb}
\usepackage{mathtools}
\usepackage{amsthm}

\DeclareMathOperator*{\argmin}{arg\,min}

\usepackage[capitalize,noabbrev]{cleveref}

\theoremstyle{plain}

\theoremstyle{definition}

\theoremstyle{remark}

\usepackage[textsize=tiny]{todonotes}

\icmltitlerunning{From Pedestrian Detection to Crosswalk Estimation: An EM Algorithm and Analysis on Diverse Datasets}

\begin{document}

\twocolumn[
\icmltitle{From Pedestrian Detection to Crosswalk Estimation: An EM Algorithm and Analysis on Diverse Datasets}



\icmlsetsymbol{equal}{*}

\begin{icmlauthorlist}
\icmlauthor{Ross Greer}{equal,yyy}
\icmlauthor{Mohan Trivedi}{equal,yyy}
\end{icmlauthorlist}

\icmlaffiliation{yyy}{Laboratory for Intelligent & Safe Automobiles, University of California, San Diego, USA}

\icmlcorrespondingauthor{Ross Greer}{regreer@ucsd.edu}

\icmlkeywords{Machine Learning, Cluster Analysis, Intelligent Transportation Systems, Pedestrian Safety, Optimization}

\vskip 0.3in
]




\begin{abstract}
In this work, we contribute an EM algorithm for estimation of corner points and linear crossing segments for both marked and unmarked pedestrian crosswalks using the detections of pedestrians from processed LiDAR point clouds or camera images. We demonstrate the algorithmic performance by analyzing three real-world datasets containing multiple periods of data collection for four-corner and two-corner intersections with marked and unmarked crosswalks. Additionally, we include a Python video tool to visualize the crossing parameter estimation, pedestrian trajectories, and phase intervals in our public source code.
\end{abstract}

\renewcommand{\topfraction}{.75}

\section{Introduction}
The detection of crosswalks serves as an important module in pedestrian detection and tracking systems for both intelligent vehicles and smart infrastructure, especially when it comes to safety, as accidents involving pedestrians is one of the leading causes of death and injury around the world \cite{gandhi2008computer}. 

The expected behavior of a vehicle (autonomous or not) in such situations is made explicitly clear by the California DMV Driver Handbook: ``If you approach a pedestrian crossing at a corner or other crosswalk, even if the crosswalk is in the middle of the block, at a corner with or without traffic signal lights, whether or not the crosswalk is marked by painted lines, you are required to exercise caution and reduce your speed, or stop if necessary, to ensure the safety of the pedestrian." 

Autonomous vehicles adhere to this policy and reduce speed when a crosswalk is detected; such control methods are described in \cite{zhu2021interactions}, also accounting for the uncertainty in occluded scenes. Driver assistance systems have also been designed to encourage similar slowing; for example, the detection of a crosswalk activates a pedestrian collision warning system in \cite{5548120}. Modules and data supporting these practices are extensive; as examples, the crowd-detection system sensitivity in \cite{reisman2004crowd} is tuned based on the presence of a crossing, and driver slowing patterns at crosswalks and associated pedestrian behaviors have been thoroughly studied in \cite{zang2021quantitative} and \cite{robert2021automated}, providing a useful model for autonomous vehicles to adopt. Further, autonomous vehicles exercise caution to ensure safety of pedestrians through the estimation of pedestrian paths, which can be enhanced by the presence of crosswalks which can be used to estimate vehicle risk of encountering a pedestrian \cite{tian2014estimation} and suggest a higher probability of pedestrian crossing behavior as well as a prescribed motion \cite{gandhi2008image}. As some motivating examples, \cite{9409688} shows that knowledge crosswalk boundaries improve this path prediction using a modified social force model, and \cite{8370119} explains that spatial layout of the environment, including crosswalks, provides valuable information to Automatic Emergency Braking Systems. \cite{8317865} and \cite{kitani2012activity} use scene information such as crosswalks, origins, and destinations to learn a prior distribution on future pedestrian trajectories, and this information appears repeatedly in the research in the literature review on pedestrian behavior prediction presented in \cite{8569415}. The principle is further evident in \cite{pmlr-v164-zhu22a}, which shows that contextual information is particularly useful in trajectory prediction when using the technique of `unlikelihood training', where trajectory outputs which violate contextual expectations are assigned lower probability. \cite{fridovich2020confidence} shows that knowledge of goals not only informs trajectory predictions, but more importantly can properly adjust an autonomous system's expectations in its prediction confidence when a pedestrian behaves in a way which deviates from direct paths to its goal; modeling this uncertainty can help autonomous agents perform more safely in human interactions.

\begin{figure*}[ht]
    \centering
    \includegraphics[width=\textwidth]{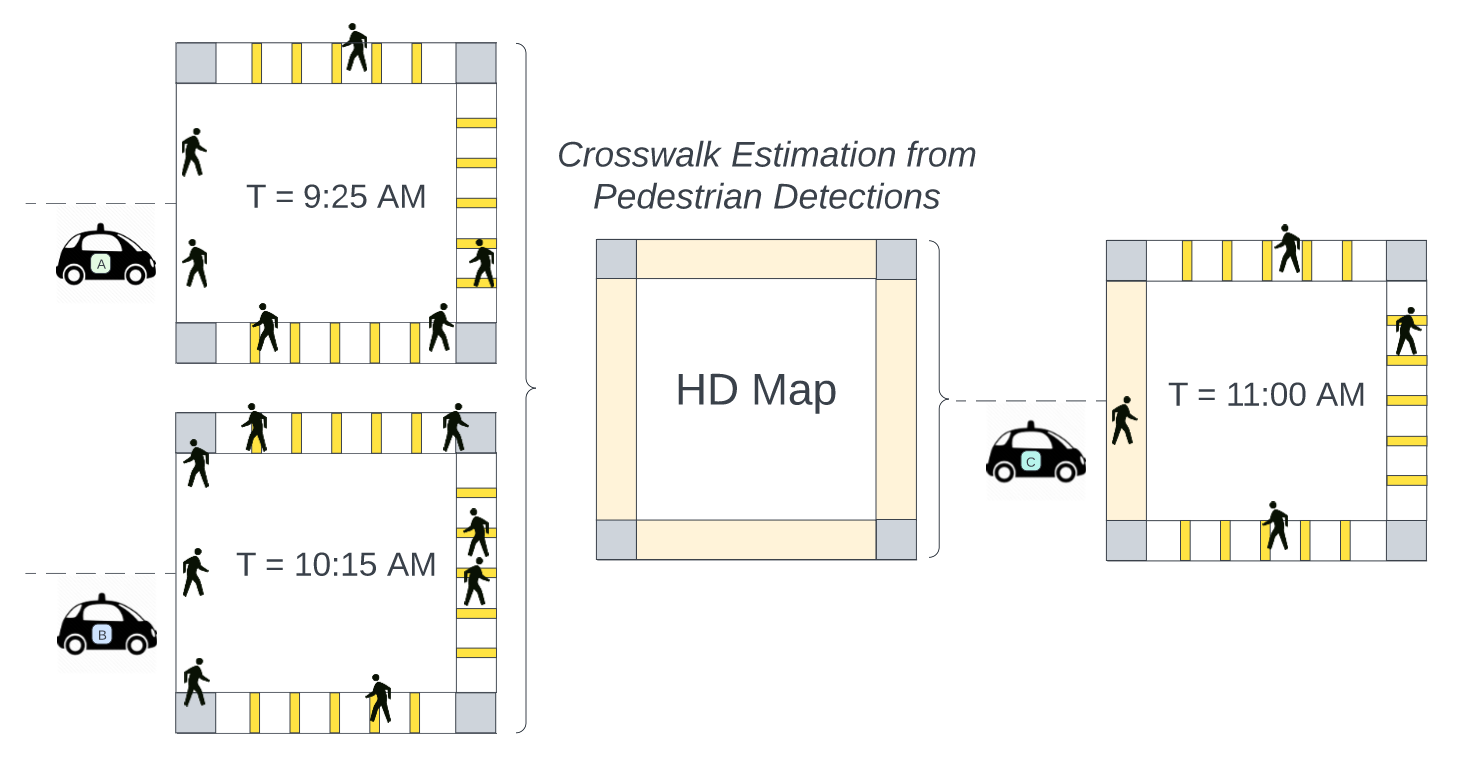}
    \caption{In this scene, two different intelligent vehicles approach the same intersection at different times, detecting pedestrians on both marked and unmarked crosswalks. The detections are provided to the crosswalk estimation algorithm proposed in this paper. The output crosswalk parameters are used to update an HD map. Intelligent vehicles approaching the same intersection now have information about the presence of the unmarked crosswalk, which can be employed in Automatic Emergency Braking Systems or similar systems for pedestrian safety and comfort, as well as systems designed to predict and classify pedestrian trajectories.}
    \label{fig:mot}
\end{figure*}

Awareness of crossings is of particular importance in high-speed environments, where street cross markings can lure pedestrians into a false sense of security when drivers are unaware of the crossing \cite{4298901}. 

In addition to benefits of pedestrian and driver safety, even the vehicle's own path planning objectives and related tasks can benefit from accurate crosswalk detections; as an example, vehicle localization can be longitudinally corrected based on crosswalk estimates \cite{6629538}.   

\begin{figure}[ht]
    \centering
    \includegraphics[width=.49\textwidth]{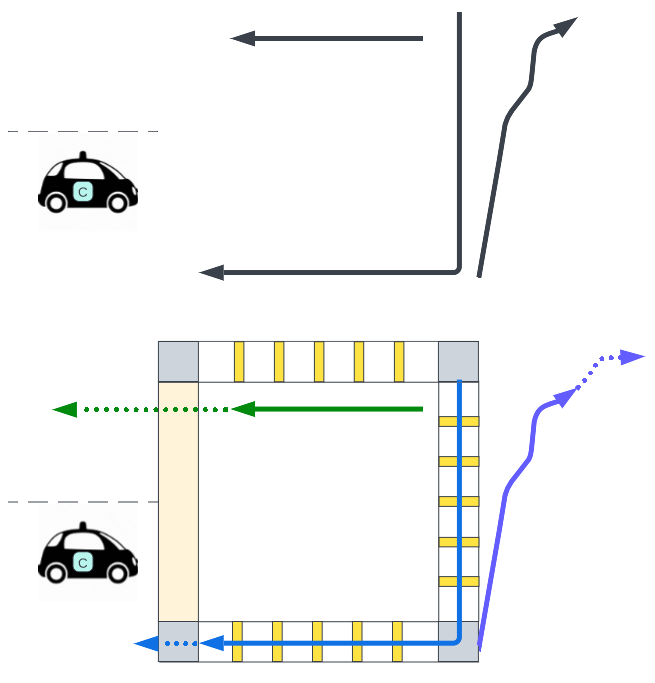}
    \caption{Understanding pedestrian scene infrastructure parameters provides stronger information for the intelligent vehicle to classify the pedestrians, which in turn informs possible future trajectories. In this illustration, the pedestrian in green is riding through a bike lane, in blue reaching a corner destination, and in purple jaywalking to a sidewalk.}
    \label{fig:imp}
\end{figure}

Therefore, we propose an algorithm which, given a set of 3D pedestrian detections, estimates lines which model the de facto pedestrian crosswalks at an intersection. Figures \ref{fig:mot} and \ref{fig:imp} illustrate the purpose of the algorithm within the context of connected safety systems, where an accurate representation of scene information can improve effectiveness of vehicle safety measures taken at crossings by inferring marked and unmarked pedestrian paths to provide stronger priors for trajectory prediction. 

\section{Related Research}

Past approaches to detecting crosswalks have been primarily based on camera images of the static scene. Such methods include figure-ground segmentation \cite{coughlan2006fast}, edge direction and cross-ratio analysis \cite{5548120}, Canny edge detection and Hough line parameter estimation \cite{9481094}, MSER and ERANSAC \cite{7313537}, laser scanning \cite{7281958}, monocular image deep learning to classify presence of crosswalk in overall scene \cite{TUMEN2020123510}, PCA-based road segmentation and crosswalk detection from a LiDAR point cloud \cite{riveiro2015automatic}, and image classification using deep learning over massive instances from satellite imagery \cite{berriel2017automatic}.  

Camera-based approaches are the most popular for such detections, and for good reason: \cite{SONG2022159} shows that in urban point cloud maps, crosswalks and other road surface markings (RSMs) are the most challenging class in point cloud semantic segmentation. But even in camera-based approaches, cracks and other damage to painted lines can provide significant challenges, with remedies proposed in \cite{ito2021detection}.  

Rather than modeling the scene based using computer vision applied to static scene elements, we are the first to use the motion patterns of detected pedestrians to estimate the crosswalks. Our approach provides a few unique benefits:
\begin{itemize}
    \item The algorithm can directly recognizes crosswalks even under the additional DMV stipulation ``whether or not the crosswalk is marked by painted lines". Pedestrians following such trajectories are not necessarily illegally jaywalking; they are simply using an unmarked crosswalk. 
    \item The algorithm parameterizes crosswalks based on actual crossing paths and rest points, which may differ from those indicated by prescribed road markings.
    \item No prior knowledge of the intersection geometry is required, only pedestrian detections.
    \item The algorithm is robust to non-square, non-rectangular, and even non-parallelogram interesection-crossing geometries. 
    \item The algorithm is robust to natural degradation of road markings (chipped paint, graffiti) or temporary detours to paths (e.g. construction).
\end{itemize}

\section{EM Algorithm for Crosswalk Estimation}

We make the following assumptions about pedestrian behavior:
\begin{enumerate}
    \item The crosswalk being utilized by a pedestrian $p_i$ is a latent variable $z_{p_i}$ on which their motion depends.
    \item Pedestrians spend, on aggregate, more time at a crossing corner than on any point of the street crossing. This enables the use of k-means clusters for initialization of crossing corners. 
    \item Crosswalks are linear in nature. This enables the estimation of crossings as lines of best-fit. 
    \item Most pedestrians cross within the designated or de facto crossing bounds with an average trajectory along the middle of the crosswalk. This enables the filtering of outlier crossing points.
\end{enumerate}

\begin{figure*}
    \centering
    \includegraphics[width=\textwidth]{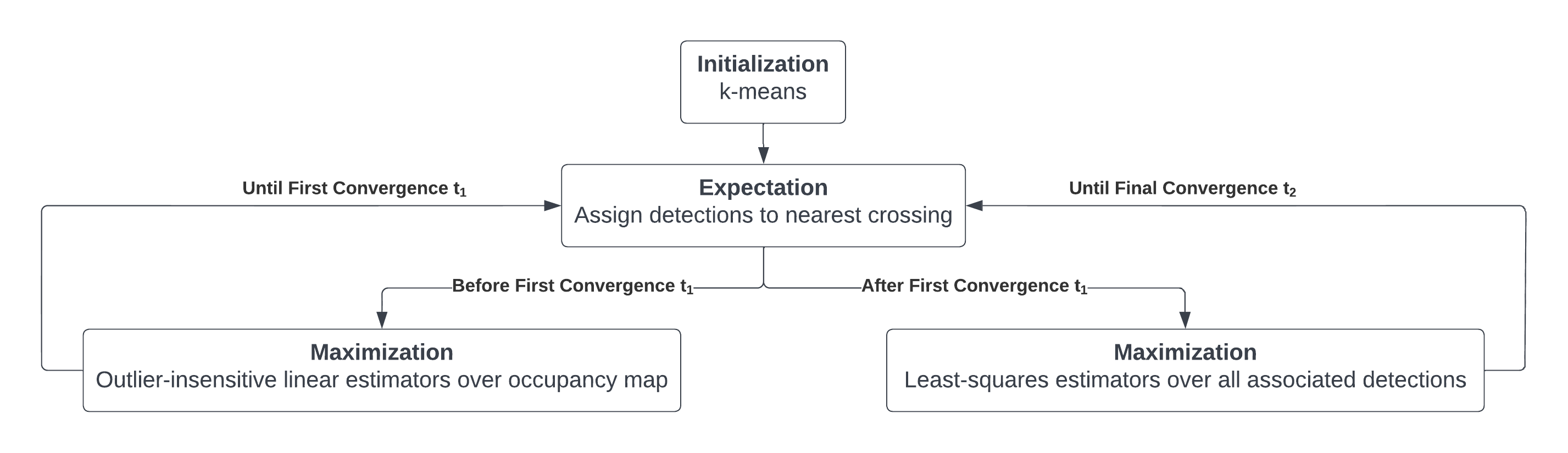}
    \caption{Proposed Algorithm for Crosswalk Estimation}
    \label{fig:emflow}
\end{figure*}

The Expectation-Maximization (EM) algorithm \cite{dempster1977maximum} is an iterative algorithm used to find maximum-likelihood estimates of parameters of a statistical model in the presence of latent (hidden) variables. We frame our statistical model to be the distribution of pedestrian detections, subject to the parameters which define the location of street corners. These corners are parameters which dually define the crosswalks as the connecting line segments. Samples from the distribution of the pedestrian detections depends on a latent variable: which crosswalk is the detected pedestrian occupying? Given input of 3D pedestrian detections within a coordinate system, the following EM algorithm  estimates the locations of street corners which act as starting or ending points for crosswalks (marked or unmarked). An overview of the algorithm is shown in Figure \ref{fig:emflow}.

\begin{enumerate}
    \item \textit{Initialization}
    \begin{enumerate}
        \item Begin with $(x,y)$ ground plane positions for all pedestrian detections made using LiDAR points.
        \item Perform k-means clustering \cite{likas2003global} where $k$ is the number of corners expected at the intersection. In the case of our dataset, $k=4$. These cluster centers will serve as preliminary corners. 
        \item Compute the angle between each pair of corners. 
        \item Using the corners central to the greatest four angles, form four pairs $(c_n, c_m)$ of corners, which parameterize a given crossing segment. The greatest angles are used to ensure that these segments comprise the border of the intersection rather than diagonal segments which interfere with traffic. In the case of environments where diagonal crossing is expected, additional segments can be allowed. 
    \end{enumerate}
    \item \textit{Expectation} 
        \begin{enumerate}
        \item Assign to each pedestrian detection $p_i$ an expected value of the latent variable $z_{p_i}$ referring to the associated crossing line $l_j$, parameterized as $(a_j, b_j, c_j)$. In this case, we compute this nearest crosswalk value using 
        \begin{equation}
            z_{p_i} = \argmin_j{\frac{|a_j x_{p_i} + b_j y_{p_i} + c_j|}{\sqrt{a_j^2 + b_j^2}}}
        \end{equation}
        \item Filter by ignoring outlier points which are more than $M$ meters from all crossing lines during this EM iteration. $M$ is a hyperparameter; we use a value of $3.5$ meters. 
        \end{enumerate}
    \item \textit{Maximization}
        \begin{enumerate}
            \item Fit a line of least-squares $l_{z_{p_i}}$ to each set of detections associated with a unique $z_{p_i}$, representing the crossing. It is important to note that initial estimates are easily trapped in local minima (for example, if k-means initializes near a non-corner position that a pedestrian chose to stand in wait). To allow the estimate to escape this local well, we use two remedies: 
            \begin{enumerate}
                \item Treat the detections as an occupancy map rather than a heatmap (that is, each location within some resolution can be either occupied or non-occupied, rather than allowing multiple detections to be representative of the same area). This is useful in reducing the effect of a person standing in a non-corner area for a long period of time, while still capitalizing on the density of points near corners from regular pedestrian traffic. We use a resolution of 0.1 meters.
                \item Begin the estimate with a line estimate method which is less sensitive to outliers (in our case, we use the Theil-Sen Estimator \cite{scikit-learn}). 
            \end{enumerate}
            Both of these modifications can be dropped once the intersection corners make an initial convergence within some tolerance $t_1$. 
            \item Define two parallel boundary lines surrounding each crossing, with distance offset 
            \begin{equation}
                d = \max(d_u-\frac{n}{r_m}, d_l),
            \end{equation}
           where $n$ is the iteration number, $d_u$ and $d_l$ are the upper and lower limits to the margin of allowable pedestrian points to be considered `on' the nearest crosswalk, and $r_m$ is the rate at which these margins narrow, in meters per iteration. To find these lines, first compute the angle of inclination $\alpha = \arctan{m}$, where $m$ is the slope. The complement of $\alpha$ subtends a right triangle whose opposite side is $d$ and hypotenuse is $b'$, a term which can be added to the initial y-intercept of the line to form a parallel line offset at distance $d$. Note that this step should be skipped until $t_1$ is satisfied; else the restrictive margins may `race' against the initial convergence of corners to a space away from outliers. 
            \item Filter points outside the boundaries from the next iteration. As $n$ increases, the boundaries tighten to include only the most representative points of the crossing. This also reduces the effects of trajectories deviating from the crossing lines after the pedestrian's segment between the corners is complete. 
            \item Repeat \textit{E} and \textit{M} steps until convergence. To determine stopping criteria, take the intersection of each pair of lines as the corresponding crossing corner, and terminate when the corner positions do not change by more than tolerance $t_2$ in sum. Both $t_1$ and $t_2$ are hyperparameters; we use values of $0.3$ and $0.05$ meters. 
        \end{enumerate}
\end{enumerate}

\section{Analysis}
\subsection{Datasets}

\begin{figure}
    \centering
    \includegraphics[width=0.4\textwidth]{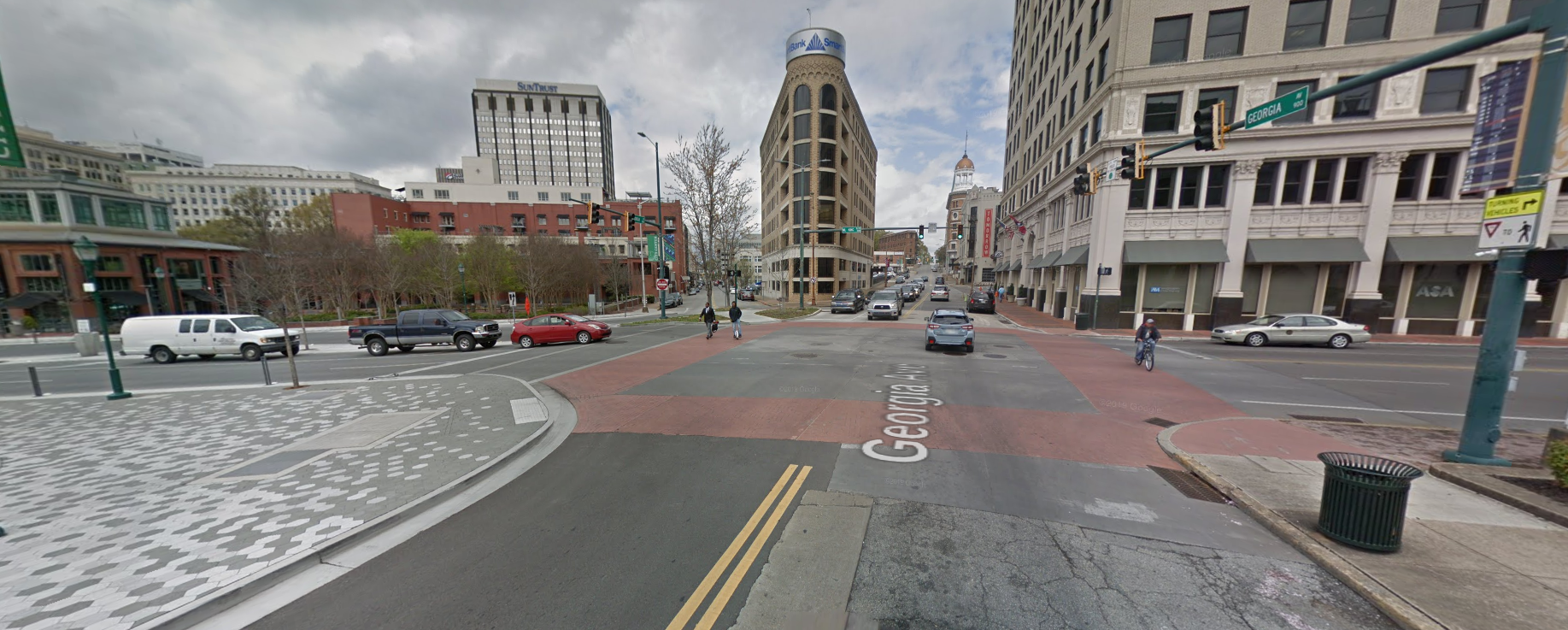}
    \includegraphics[width=0.4\textwidth]{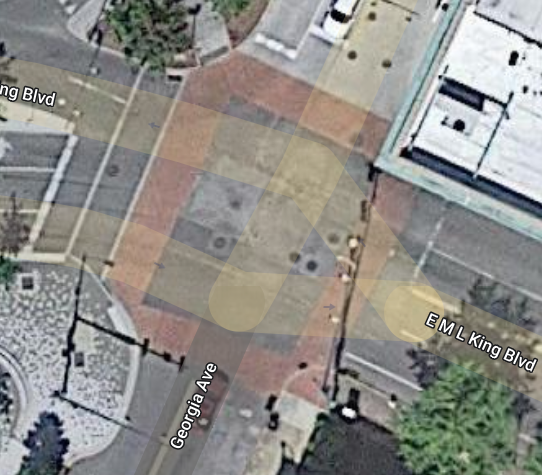}
    \caption{Ground and bird's-eye-view of the intersection of MLK and Georgia Avenue in Chattanooga, Tennessee, where infrastructure-mounted LiDAR pedestrian detections were captured.}
    \label{fig:chattanooga}
\end{figure}

\begin{figure}[ht]
    \centering
    \includegraphics[width=0.4\textwidth]{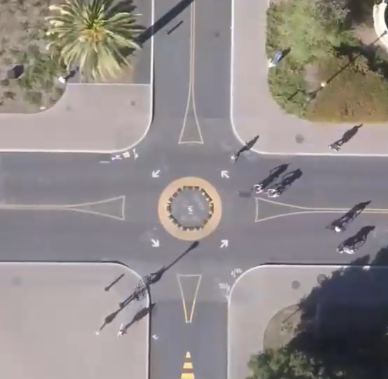}
    \caption{Bird's-eye-view of the traffic circle in Palo Alto, California, from the Stanford Drone Dataset. It is notable that this intersection contains no marked crosswalks.}
    \label{fig:furd}
\end{figure}

\begin{figure}[ht]
    \centering
    \includegraphics[width=0.35\textwidth]{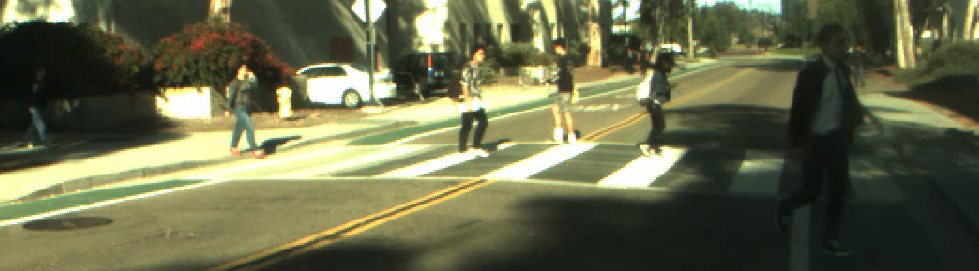}
    \includegraphics[width=0.35\textwidth]{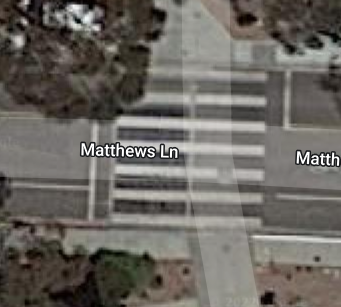}
    \caption{Ground and bird's-eye-view of the crosswalk in La Jolla, California, where vehicle-mounted camera pedestrian detections were captured.}
    \label{fig:ucsd}
    \vspace{-.5cm}
\end{figure}

Our experiments utilize three datasets. First, we utilize the data presented in the 2022 Transportation Review Board Committee on Artificial Intelligence and Advanced Computing Applications Transportation Forecasting Competition \cite{trbdatachallenge}. Three time periods of data were collected from three Ouster OS1-128 LiDAR sensors at the MLK and Georgia Avenue intersection in Chattanooga, Tennessee (see Figure \ref{fig:chattanooga}). The LiDAR point cloud data was processed by Seoul Robotics software to provide object labels (pedestrian, vehicle, and bicyclist), and sub-classification features of object size, position, and velocity. Each pedestrian data entry also includes a confidence and tracking status. Second, we use pedestrian detection data collected from the Stanford Drone Dataset \cite{robicquet2016learning} at a bicycle traffic circle \ref{fig:furd}, where pedestrians exhibit similar behavior to street crossing. Note that this dataset provides detections on a pixel scale rather than meters. Third, we use 3D pedestrian detection data collected from a two-corner crosswalk on the campus of University of California, San Diego, collected from a camera mounted to a stationary parked vehicle (see Figure \ref{fig:ucsd}) \cite{8317865}.

\begin{figure}
    \centering
    \includegraphics[width=0.4\textwidth]{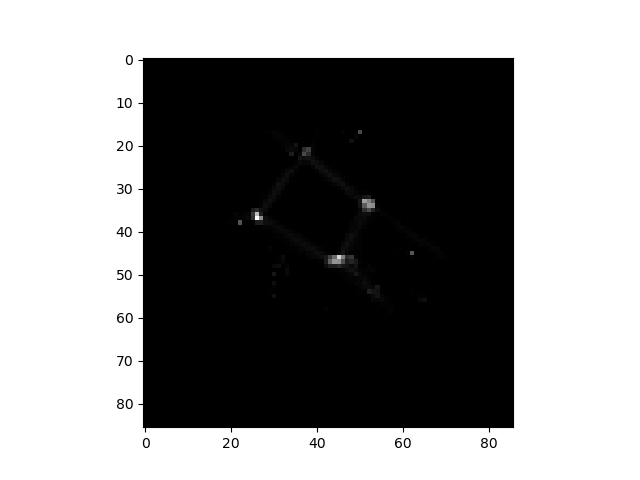}
    \caption{A heatmap visualizing pedestrian detection instances on a grid over the Chattanooga intersection. Brighter pixels indicate more pedestrian data instances belonging to that region. From the figure, we can see that the most points occur at the four corners, where a pedestrian would be waiting to cross, with faint lines over the pedestrian crosswalks.}
    \label{fig:heatmap}
    \vspace{-.25cm}
\end{figure}

The assumption that pedestrian detections tend to aggregate at crossing corners is valid for the analyzed datasets. This property is illustrated in the data collected at the intersection in Chattanooga, shown in Figure \ref{fig:heatmap}. 

\subsection{Results}

\subsubsection{Four-Corner Crossing: Chattanooga}
\begin{figure}[ht]
    \centering

    \includegraphics[width=0.45\textwidth]{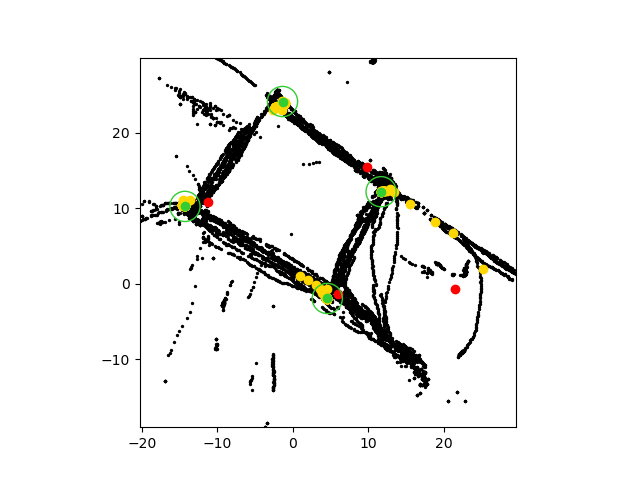}
    \caption{Estimation of street corners associated with crosswalks for the first batch of data collected in Chattanooga. Pedestrian positions are represented in black, initial cluster centers estimating corners in red, progression of the corners between iterations in yellow, and final corner positions in green.}
    \label{fig:ped_scatter1}
\end{figure}

\begin{figure}[ht!]
    \centering
    \includegraphics[width=0.45\textwidth]{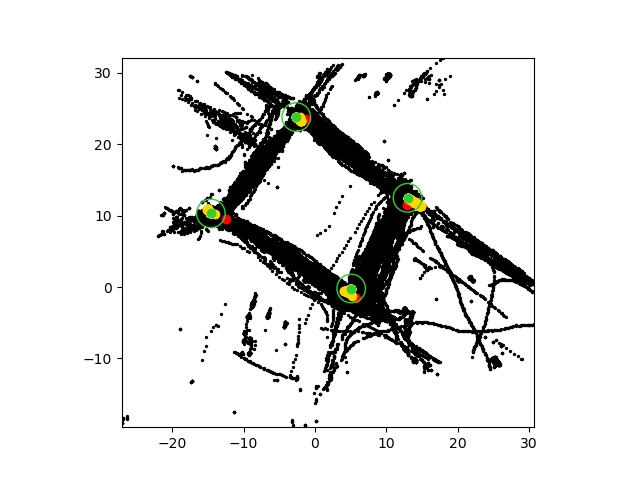}
    \caption{Estimation of street corners associated with crosswalks for the second batch of data collected in Chattanooga, following the same color scheme as described in Figure \ref{fig:ped_scatter1}.}
    \label{fig:ped_scatter2}
\end{figure}

\begin{figure}[ht]
    \centering
    \includegraphics[width=0.45\textwidth]{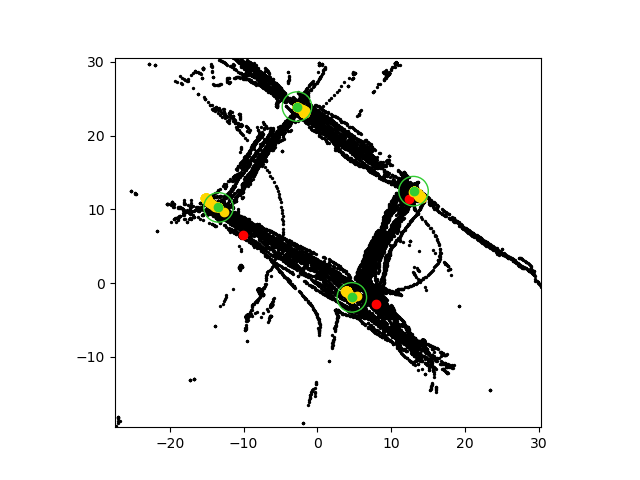}
    \caption{Estimation of street corners associated with crosswalks for the second batch of data collected in Chattanooga, following the same color scheme as described in Figure \ref{fig:ped_scatter1}.}
    \label{fig:ped_scatter3}
\end{figure}

Figures \ref{fig:ped_scatter1}, \ref{fig:ped_scatter2}, and \ref{fig:ped_scatter3} illustrate the application of the above EM algorithm to estimate the location of crossing corners for each of the three data collection periods. Below each illustration is a graph which shows the decrease in residuals from the least-squares line (averaged over the four clusters) for each iteration of the algorithm. Observing the coverage of the pedestrian detections, it is evident that the final EM algorithm estimates (red) are much closer to the true crosswalk corners than the initial k-means estimates (cyan). Notably, the EM algorithm is able to move the corner estimates to areas less densely populated with detection points to improve accurate positioning, and is robust to the many noisy data points which exist in non-crossing regions. Despite the three samples containing completely different detection datapoints and k-means initializations, the corners converge to similar locations as described in Table \ref{table:1}, with an average difference of 0.667 meters from the estimate based on the aggregate dataset.

\begin{table}[h!]
\centering
\caption{Distance (m) of batch estimate from aggregate corner estimate}

\begin{tabular}[width=0.5\textwidth]{c|c|c|c|c|c}
     \small Corner: & \small 1 &\small 2 & \small 3 & \small 4 &  \small Average \\
      \hline
   \small Batch 1 & 1.282 & 0.227 & 1.615 & 1.273 & 1.111 \\  
   \small Batch 2 & 0.030 & 0.085 & 0.223 & 0.112 & 0.112 \\
   \small Batch 3 & 0.217 & 1.161 & 1.553 & 0.181 & 0.778 \\
\end{tabular}
\label{table:1}
\end{table}

The first batch (Figure \ref{fig:ped_scatter1}) shows the most dramatic shift from initial to final locations. This data period was heavily skewed by pedestrian behavior which occurred down the road from the intersection, but the algorithm was able to overcome the initial local minimum through the important step of occupancy-map formulation and outlier-insensitive linear estimation.

\begin{figure}[ht]
    \centering
    \includegraphics[width=0.4\textwidth]{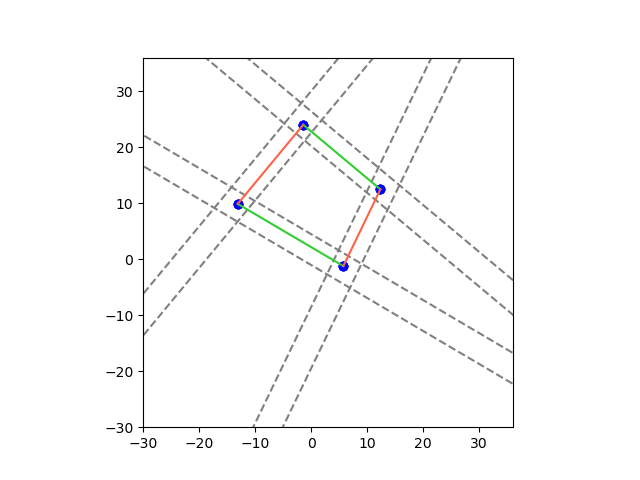}
    \caption{Model representation of the intersection. Corner points are depicted in blue, representing the average of the corners found across the three data batches. In this figure, to best illustrate the scene, parallel crossings are shown as solid green lines due to an active crossing phase, while the other set of parallel crossings are shown in red to indicate non-active crossing phase. Dashed lines indicate our defined inner and outer boundaries for each crossing.}
    \label{fig:crossings}
\end{figure}

The corresponding crosswalks are illustrated in Figure \ref{fig:crossings}, including a margin inside and outside each line to account for pedestrian deviations from the center crossing line. Depending on intended application, this margin can be provided as a hyperparameter to match safety or legal expectations, or learned from the data to match practical usage. For example, in our illustration, we use values for inner and outer margins of $I=2$ and $O=2.75$ respectively. The greater tolerance for $O$ is employed since pedestrians can safely wait further from the corner than the sidewalk, but cannot safely wait within the street.

\begin{figure}
    \centering
    \includegraphics[width=0.5\textwidth]{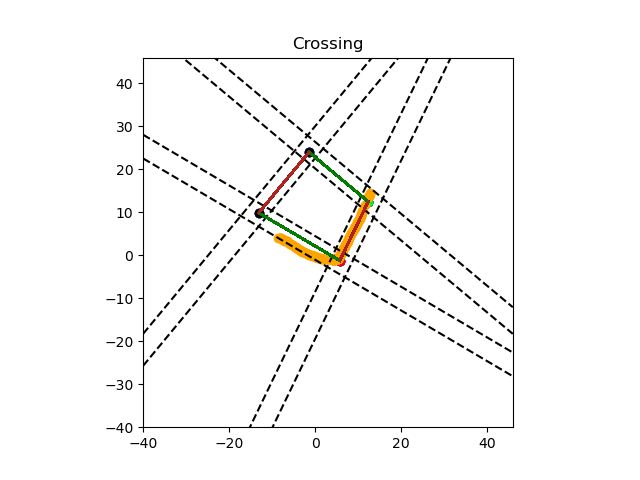}
    \includegraphics[width=0.5\textwidth]{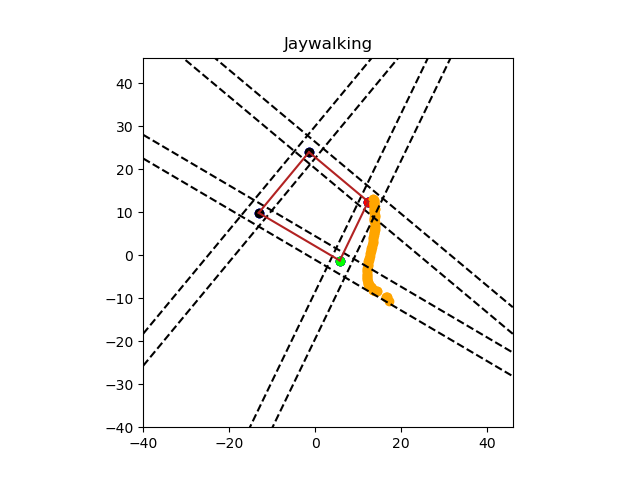}
    \includegraphics[width=0.5\textwidth]{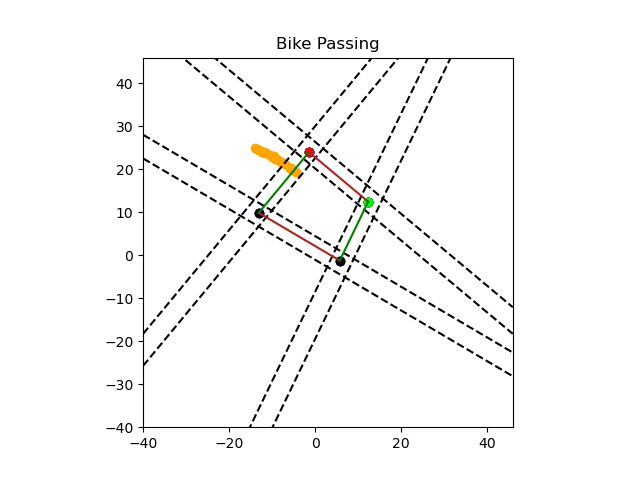}
    \caption{With information on the location of street corners and associated crosswalks, rudimentary rule-based classifiers can distinguish between trajectories of classes such as standard crossing behavior, jaywalking, and bikes passing through on the road. Understanding pedestrian dynamics is crucial to trajectory estimation for safe autonomous vehicle control.}
    \label{fig:class}
\end{figure}

Using the estimated corners and crosswalks, we show that it is possible to classify pedestrian trajectories using a rule-based technique (i.e. pedestrians passing travelling from corner-to-corner are legally crossing, pedestrians who move through an intersection via points adjacent to but not upon corners are likely bikers, pedestrians who cross from a sidewalk to another sidewalk but avoid the corners are jaywalking, etc.). These classifications, shown in Figure \ref{fig:class}, have value in predicting future trajectories of the pedestrians for safe vehicle operation.  

\begin{figure}[ht]
    \centering
    \includegraphics[width=0.45\textwidth]{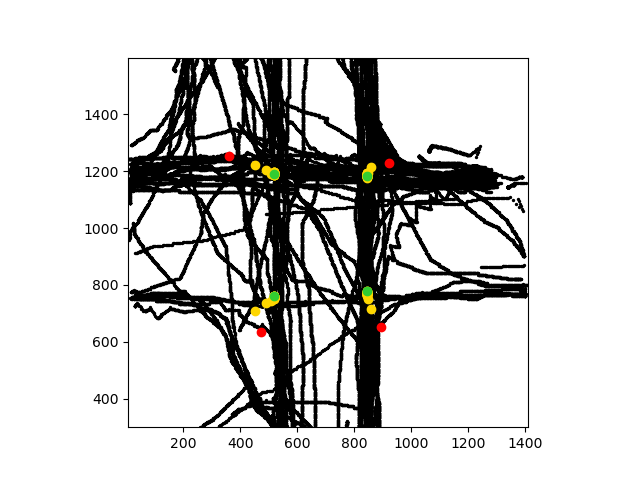}
    \caption{Estimation of crossing corners associated with crosswalks for the traffic circle in the Stanford Drone Dataset, following the same color scheme as described in Figure \ref{fig:ped_scatter1}.}
    \label{fig:pedfurd}
\end{figure}

\begin{figure}[ht]
    \centering
    \includegraphics[width=0.45\textwidth]{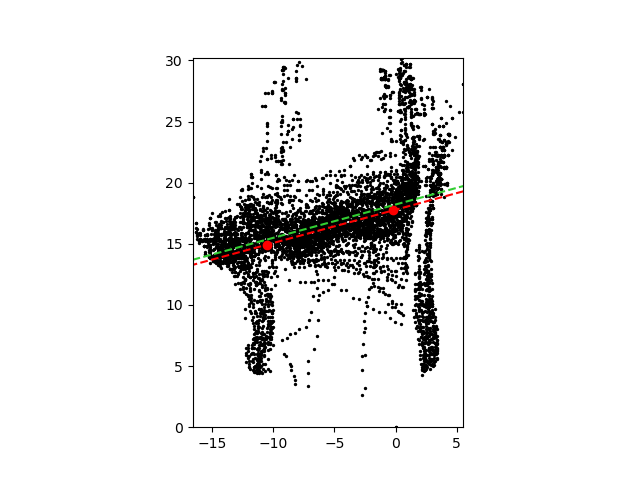}
    \caption{Model representation of the La Jolla crosswalk. Initial estimate using k-means in red, with the final line following iterative refinement shown in green.}
    \label{fig:lisawalk}
\end{figure}

\subsubsection{Four-Corner Crossing: Stanford Traffic Circle} 

We repeat the proposed algorithm on a bike traffic circle from the Stanford Drone dataset. Most notably, this consists entirely of unmarked crosswalks, yet the pedestrian trajectories still produce a clear pattern which is otherwise invisible in road infrastructure. The k-means initialization places the estimates in locations which are near the corners but not precise; after 27 iterations, the corners converge to positions which represent the most frequent (unmarked) crosswalks used by the pedestrians.

\subsubsection{Two-Corner Crossing: La Jolla}

Using the modification of the algorithm for a single crosswalk (no intersecting corners), we demonstrate the algorithm's effectiveness at locating the two crossing corners at the La Jolla crosswalk. The initial estimate reflects the noise of those walking on the pathway (notice the density in the lower half of the scatter plot in Figure \ref{fig:lisawalk}), and after refining the estimate by narrowing the accepted margin of inliers, the crossing better reflects the tendency of pedestrians to gather and orient their trajectory diagonally upward within our bird's-eye-view. Notably, this is not the perpendicular crossing geometry marked by the road, but rather a de facto crossing pattern.

\section{Concluding Remarks}
In this work, we described and illustrated the effectiveness of an EM algorithm for estimating crosswalks using a real-world pedestrian detection dataset. This algorithm can be used to model both marked and unmarked crossings in a variety of geometries. Crosswalk detection methods which rely on the physical appearance of road infrastructure naturally fail in cases where the infrastructure is damaged, obstructed, or ignored; by contrast, methods which make use of pedestrian motion can still provide reliable information under these conditions. 
\subsection{Limitations}
While the techniques introduced in the two-stage EM algorithm are designed to make the algorithm robust to certain violations of the prerequisite assumptions, the algorithm is less effective in situations where pedestrians congregate at points within the considered area but away from the crossing origins or destinations. In such cases, the k-means clustering provides a poor initialization (for example, if there are two congregating points at one corner, and a severe imbalance of data at another corner, then two cluster centers will share the same corner and may not converge to the crossing geometry). It would therefore be important to add appropriate check conditions to be sure initial corner estimates are sufficiently spaced. Second, though the best-representative crossing line is determined, further investigation to estimate the effective crossing margin would be valuable, as this is currently passed to the algorithm as a hyperparameter and may vary in different crossing scenes. 
\subsection{Broader Impact}
The demonstrated datasets were collected from both city infrastructure LiDAR and vehicle-mounted camera; any modality which can perform 3D detection is suitable. This method readily extends to measurements made by moving vehicles with accurate localization, and can be used downstream to support safe interactions between pedestrians and autonomous vehicles. As explained in \cite{gandhi2008computer}, detection and tracking of objects from cameras is easier due to absence of camera motion and use of background subtraction, but in addition to infrastructure-based sensing, improvements in vehicle-to-vehicle and vehicle-to-infrastructure communication will allow for improved exchange of temporal information or information about objects that are seen by one but not seen by others. In particular, vehicles which are able to report pedestrian detections within a global coordinate frame can supply local data to HD maps, leveraging the above algorithm to maintain an accurate, dynamic estimate of de facto crosswalks with repeated samples over time. 

\clearpage

\bibliography{biblio}
\bibliographystyle{icml2022}



\end{document}